\documentclass[letterpaper, 10 pt, conference]{ieeeconf}  

\IEEEoverridecommandlockouts                              

\usepackage{graphics} 
\usepackage{makecell}
\usepackage{cite}
\usepackage{soul}
\usepackage[noend]{algpseudocode}
\usepackage{amssymb}
\usepackage{multirow}
\usepackage{algorithm}
\usepackage{amsmath} 
\usepackage{amssymb}  
\usepackage{todonotes}
\usepackage{multicol}
\usepackage{adjustbox}

\title{\LARGE \bf
Improving Out-of-Distribution Generalization of Learned Dynamics by Learning Pseudometrics and Constraint Manifolds
}

\author{Yating Lin$^{1}$, Glen Chou$^{2}$, and Dmitry Berenson$^{1}$
\thanks{This work was supported in part by the Office of Naval Research Grant N00014-21-1-2118 and NSF grants IIS-1750489, IIS-2113401, and IIS-2220876. $^{1}$University of Michigan {\tt\small\{yatinlin, dmitryb\}@umich.edu}  $^{2}$Massachusetts Institute of Technology, {\tt\small gchou@mit.edu}}
}

\begin{document}

\definecolor{darkgreen}{rgb}{0., 0.5, 0.}

\newcommand{\nx}{n}
\newcommand{\na}{m}
\newcommand{\nc}{c}
\newcommand{\D}{\mathcal{D}}
\newcommand{\Z}{\mathcal{Z}}
\newcommand{\Dist}{P}

\maketitle
\thispagestyle{empty}
\pagestyle{empty}

\begin{abstract}

We propose a method for improving the prediction accuracy of learned robot dynamics models on out-of-distribution (OOD) states. We achieve this by leveraging two key sources of structure often present in robot dynamics: 1) sparsity, i.e., some components of the state may not affect the dynamics, and 2) physical limits on the set of possible motions, in the form of nonholonomic constraints. Crucially, we do not assume this structure is known \textit{a priori}, and instead learn it from data. We use contrastive learning to obtain a distance pseudometric that uncovers the sparsity pattern in the dynamics, and use it to reduce the input space when learning the dynamics. We then learn the unknown constraint manifold by approximating the normal space of possible motions from the data, which we use to train a Gaussian process (GP) representation of the constraint manifold. We evaluate our approach on a physical differential-drive robot and a simulated quadrotor, showing 
improved prediction accuracy on OOD data relative to baselines.


\end{abstract}

\vspace{-6pt}
\section{Introduction}

A key component of the robot autonomy stack is the dynamics model, which predicts how the robot state changes given the current state and control. Since the dynamics of many systems are difficult to hand-model, a popular choice is to learn them from data using, e.g., neural networks (NNs) \cite{DBLP:conf/icra/NagabandiKFL18} or Gaussian processes (GPs) \cite{DBLP:conf/icml/DeisenrothR11}. This learned model is then used for planning and control. A key assumption for these models to work well is that the training data is drawn in an independently and identically distributed (i.i.d.) fashion from the same distribution where the model will be deployed. This is often violated in robotics, where we may need to visit states and controls that are significantly different from the training data, referred to as out-of-distribution (OOD) inputs \cite{DBLP:journals/corr/abs-2108-13624}. In OOD domains, the model accuracy can degrade and hamper the robot's performance, revealing a critical need for learned dynamics that generalize better on OOD inputs.

To work towards this goal, we explore two insights: that many robots 1) have \textit{sparse} dynamics, i.e., not all state variables affect the dynamics, and 2) satisfy nonholonomic \textit{constraints} arising from physics and design, e.g., a car cannot translate sideways. Enforcing that our learned model conforms to this information can improve accuracy, as na\"ively-trained dynamics may predict highly non-physical behavior, especially in OOD domains. However, without copious \textit{a priori} knowledge, it is difficult to know what sparsity pattern and constraints to prescribe. Moreover, idealized sparsity and hand-coded constraints can fail to hold due to hardware imperfections, or if the robot undergoes faults.  

\begin{figure}
    \centering\vspace{-10pt}
    \includegraphics[width=\linewidth]{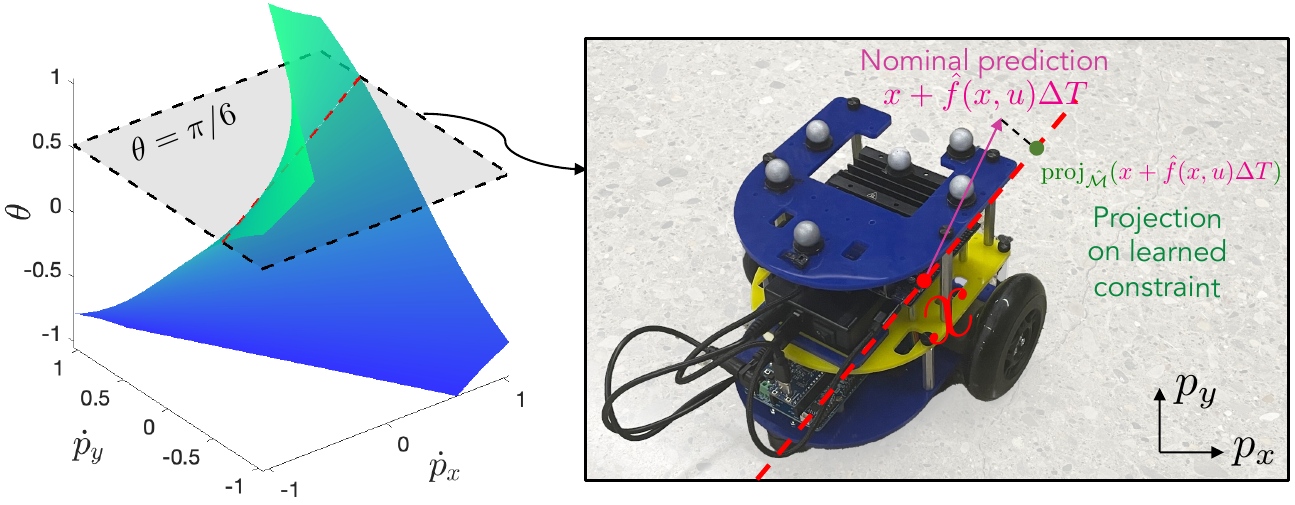}\vspace{-12pt}
    \caption{\textbf{Left}: Visualization of the constraint manifold $\mathcal{M} = \{x, \dot x \mid -\dot p_x \sin(\theta) + \dot p_y \cos(\theta) = 0\}$ satisfied by the unicycle \eqref{eqn:car}. \textbf{Right}: The unicycle dynamics are sparse, i.e., the set of possible velocities (red dotted line) is the same for all $x$ with the same orientation. We learn the sparsity pattern and constraints (on a physical differential-drive robot), and \textit{project} the predictions of our learned dynamics $\hat f(x,u)$ to be consistent with them.}
    \label{fig:manifold}
    \vspace{-15pt}
\end{figure}


In this paper, we strike a middle ground: by learning any sparsity and constraints from a small dataset, we can use them to adjust the learned dynamics to be more accurate than a na\"ive model, while also avoiding \textit{a priori} prescription of structure. In particular, we hypothesize that learned constraints can generalize better than learned dynamics in OOD domains, as they are often defined in a lower-dimensional space where data-density is easier to achieve. Our method learns a distance pseudometric that uncovers the sparsity pattern in the dynamics, which we use to perform dimensionality reduction on the input of the learned dynamics. We then identify an implicit constraint manifold in the space of state and state derivatives satisfied by feasible transitions. We restrict attention to nonholonomic constraints in this work. Finally, to make predictions, we project the output of our learned dynamics onto the learned constraint. We contribute: 
\begin{itemize}
    \item a method for contrastive learning of distance pseudometrics to identify sparsity in dynamics and constraints
    \item a framework for approximate learning of nonholonomic constraints satisfied by the robot from trajectory data
    \item a method for improving the accuracy of learned dynamics by projecting its output to satisfy learned constraints
    \item evaluation on a physical differential-drive robot and simulated quadrotor, improving accuracy over baselines
\end{itemize}

\vspace{-6pt}
\section{Related Work}
\vspace{-4pt}

Many methods that use learned dynamics for control \cite{DBLP:conf/icml/DeisenrothR11, DBLP:conf/nips/ChuaCML18} often fail far from training data. Some methods aim to mitigate this unreliability, e.g., \cite{DBLP:journals/scirobotics/MitranoMB21, DBLP:journals/ral/GuzziCNGG20} bias planners away from unreliable model transitions. Other work \cite{DBLP:journals/corr/abs-2212-06874, DBLP:journals/ral/KnuthCOB21, DBLP:conf/cdc/ChouOB21} plans safely with learned dynamics by enforcing that the system remains within a bound of the training data. This bound explicitly limits model extrapolation; in contrast, our goal is to use the model far from data, by empirically maintaining accuracy via sparsification and learned constraints.

Other methods improve OOD generalization of learned dynamics via symmetry and physical constraints. For instance, rotational symmetry of visual dynamics can be encoded via data augmentation (e.g., random crops) \cite{DBLP:conf/icml/LaskinSA20} or equivariant networks \cite{DBLP:conf/corl/WangWZP21}. Other methods improve generalization with respect to task-irrelevant features of the input data (e.g., \cite{DBLP:conf/iclr/0001MCGL21}, \cite{DBLP:conf/rss/PacelliM20}). While similar in motivation, these methods exploit the structure in image observations. Instead, our method directly learns structure in the dynamics via a sparsity pattern and constraints, complementing the above methods. Physics-informed learning has shown that enforcing Lagrange and Hamilton's equations can improve generalization in learning unconstrained \cite{DBLP:conf/iclr/LutterRP19, DBLP:conf/iclr/ZhongDC20, DBLP:conf/nips/GreydanusDY19} and constrained dynamics \cite{DBLP:conf/l4dc/GeistT20, DBLP:conf/corl/RathGT21, DBLP:conf/l4dc/DjeumouNGPT22}. In the latter work, the functional form of the constraint is assumed to be known \textit{a priori}, with parametric uncertainty. Our work sits between physics-constrained and unstructured learning, in that we discover constraints and sparsity directly from data, without assuming their existence or form \textit{a priori}. 


Finally, our work relates to constraint learning. \cite{DBLP:conf/corl/SutantoFERS20} learns equality constraints, but is limited to holonomic constraints. Other methods \cite{DBLP:conf/wafr/ChouBO18, DBLP:conf/corl/ChouOB20, DBLP:journals/corr/abs-2112-04612, DBLP:journals/tcst/MennerWZ21, DBLP:conf/icra/StockingMMT22} learn inequality constraints; but, unlike ours, require near-optimal demonstrations, and cannot learn from unstructured trajectory data, which is critical for dynamics-learning to cheaply improve data-density. The most similar method \cite{DBLP:journals/corr/abs-2112-04612} uses GPs to learn nonlinear inequality constraints, but supervises the GP via constraint gradient data that is hard to obtain for dynamics-learning. Our focus also differs: instead of safety, we explore how learned constraints can empirically improve dynamics prediction.





\section{Preliminaries and Problem Statement}\label{sec:prelims}



\noindent\textbf{Definitions}: In this paper, we consider deterministic systems
\begin{equation}\label{eqn:dyn}
    \dot{x} = f(x,u),
\end{equation}
where $f: \mathcal{X} \times \mathcal{U} \rightarrow \mathcal{X}$ and $\mathcal{X} \subseteq \mathbb{R}^\nx$ and $\mathcal{U} \subseteq \mathbb{R}^\na$ are the state and control space. The trajectories of \eqref{eqn:dyn} may implicitly satisfy constraints, e.g., a car cannot instantaneously move sideways. 
In this paper, we consider vector-valued \textit{nonholonomic equality} constraints, i.e., $\mathbf{g}(x, \dot x) = \mathbf{0}$, which are implied by \eqref{eqn:dyn}. Here, $\mathbf{g}: \mathcal{X} \times T_x(\mathcal{X})\rightarrow \mathbb{R}^\nc$,  where $\nc$ is the number of constraints and $T_x(\mathcal{X})$ is an $\nx$-dimensional vector space at every $x$. We also assume the implicit constraint $\mathbf{g}(x,\dot{x})=\mathbf{0}$ can be approximated as an affine function of $\dot{x}$:
\begin{equation}\label{eqn:pfaffian_matrix}\small
    G(x)\dot{x} + g(x) = 0 \Longleftrightarrow \underbrace{\begin{bmatrix}G(x) & g(x)\end{bmatrix}}_{\doteq \Gamma(x)}
    \begin{bmatrix}
        \dot x \\ 1
    \end{bmatrix} = 0,
\end{equation}\vspace{-24pt}

\noindent where 

\vspace{-15pt}
\begin{equation}\label{eqn:G}\footnotesize
   \hspace{-7pt}G({x}) = \begin{bmatrix} 
     g^{(1)}_1({x}) & \cdots & g^{(1)}_n({x})  \\
    & \vdots & \\
    g^{(\nc)}_1({x}) & \cdots & g^{(\nc)}_n({x}) 
\end{bmatrix};\  g(x) = \begin{bmatrix}g_0^{(1)}(x)\\\vdots\\ g_0^{(\nc)}(x)\end{bmatrix}.
\end{equation}
Constraints of the form \eqref{eqn:pfaffian_matrix} generalize Pfaffian constraints \cite{lavalle2006planning}, which omit the bias terms $g^{(i)}_0(x)$, and includes a class of nonholonomic constraints \cite{lavalle2006planning}. These implicit constraints define a feasible manifold in the space of states/derivatives

{\begin{equation}\label{eqn:manifold}
    \mathcal{M} \doteq \{(x,\dot{x})\in \mathcal{X} \times T_x(\mathcal{X}) \mid \mathbf{g}(x,\dot{x})=\mathbf{0}\}.
\end{equation}}
At a fixed $x$, we define $T_x \mathcal{M}$, of dimension $\nx - \nc$, as the set of admissible derivatives $\{\dot x \mid \exists u \in \mathcal{U}, f(x, u) = \dot x\}$. Similarly, we define the normal space $N_x\mathcal{M}$ \cite{riemann}, of dimension $\nc$, as $\{ \dot x \mid \neg\exists u \in \mathcal{U}, f(x, u) = \dot x\}$. 

\noindent\textbf{Gaussian processes}: 
In this paper, we use Gaussian processes (GPs) \cite{gpml} for dynamics learning (though our method is also compatible with NNs or other function approximators). We give a brief overview of GPs below. 
Consider a training set $D=\{\textbf{x}_i,y_i\}_{i=1}^N$, where $\textbf{x} \in \mathbb{R}^d$ and $y \in \mathbb{R}$ are the training inputs and labels, and labels $y$ are corrupted with additive Gaussian noise $\mathcal{N}(0,\sigma^2)$. Given test inputs $\textbf{x}_*$, we can infer the posterior distribution of test labels using a GP (see \cite{gpml} for details).



This process relies on a positive-definite kernel $k(\textbf{x},\textbf{x})$ with hyper-parameters $\Theta = \{\sigma_1, l_1,...,\sigma_d,l_d\}$. Here, $\sigma_i$, $l_i$ are the variances and length-scales of dimension $i$, which can be learned by maximizing the log marginal likelihood of $D$. For multi-dimensional labels $y$, we model each output dimension independently with a scalar-output GP, which we refer to as independent GPs (IGPs).




\begin{figure}
    \centering
    \includegraphics[width=\linewidth]{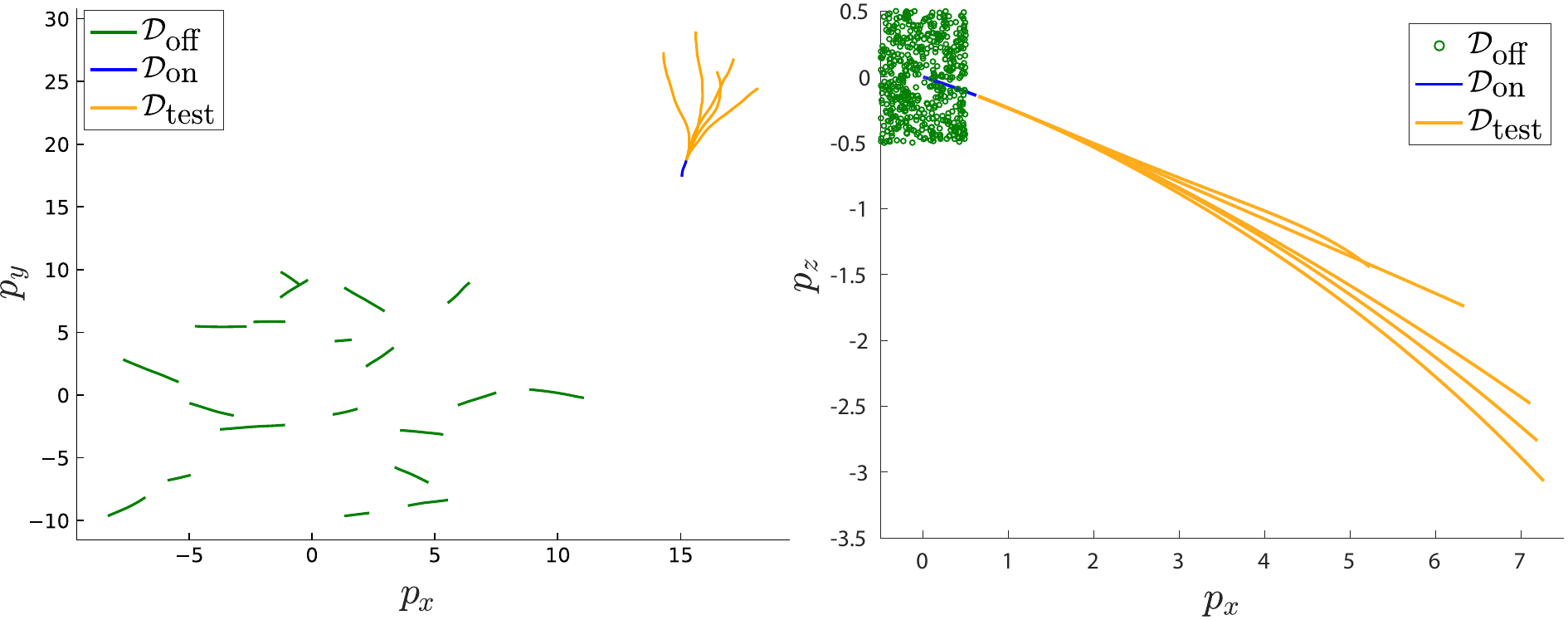}\vspace{-12pt}
    \caption{Example datasets (offline $\D_\textrm{off}$, one instance of noiseless online $\D_\textrm{on}$, and the associated test $\D_\textrm{test}$) used for training and evaluation for the unicycle \eqref{eqn:car} (\textbf{left}) and quadrotor \eqref{eqn:quadrotor} (\textbf{right}). The test data $\D_\textrm{test}$ is far from the training data $\D$, and can be considered OOD w.r.t. that data (this is confirmed by decreased prediction accuracy on these inputs; see Sec. \ref{sec:results}).}
    \label{fig:data_unicycle}
    \vspace{-15pt}
\end{figure}


\noindent\textbf{Problem statement}:
We assume the dynamics \eqref{eqn:dyn} and constraint \eqref{eqn:manifold} are unknown, but that we are given a dataset $\D$ generated by the system. We divide the dataset $\D$ into offline (online) datasets $\D_\textrm{off}$ ($\D_\textrm{on}$). $\D_\textrm{off}=\{\xi_{xu}^{(n)},\dot\xi^{(n)}\}_{n=1}^N$ contains $N$ state-control trajectories $\xi_{xu}^{(n)} \doteq (x_1^{(n)}, u_1^{(n)}, \ldots, x_T^{(n)})$, each sampled at $T$ time indices, and corresponding derivatives $\dot \xi^{(n)} \doteq (\dot x_1^{(n)}, \ldots, \dot x_T^{(n)})$ collected offline. 
During online deployment, we execute one control trajectory $u_\textrm{on}(t)$ and collect the data observed online at regular time instants, which constitutes the online dataset $\D_\textrm{on}=(\xi_{xu}^{\textrm{on}},\dot\xi^{\textrm{on}})$.


Using $\D$, we train a dynamics model $\dot x = \hat f(x,u)$. Inputs far from $\D$ are referred to as out of distribution (OOD). Our goal is to improve prediction accuracy on states that are OOD with respect to $\D$; we refer to these inputs as $\D_\textrm{test}$. See Fig. \ref{fig:data_unicycle} for examples of $\D_\textrm{off}$, $\D_\textrm{on}$, and $\D_\textrm{test}$ used in Sec. \ref{sec:results}.

\section{Method}\label{sec:method}
\begin{figure}
    \centering
    \includegraphics[width=\linewidth]{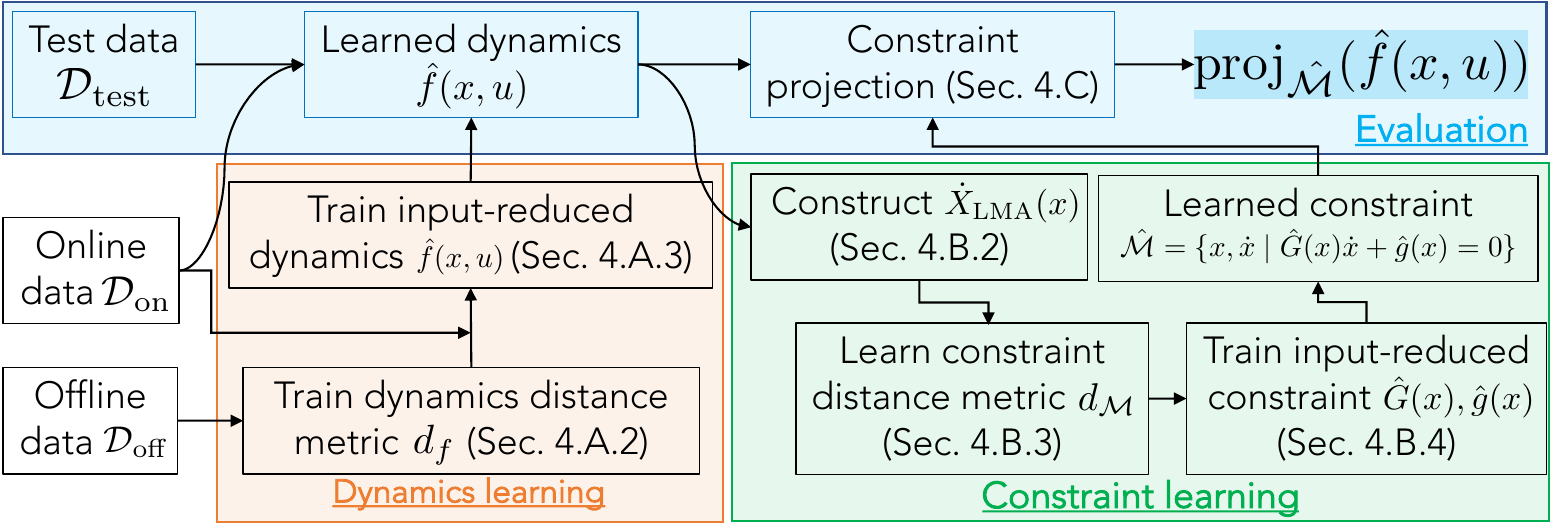}\vspace{-10pt}
    \caption{Method. \textbf{\textcolor{orange}{Dynamics learning}}: Given offline data $\D_\textrm{off}$, we learn a dynamics pseudometric $d_f$, which we use to reduce the input space dimension of the dynamics model $\hat f(x,u)$. We train $\hat f(x,u)$ on the input-reduced offline data. \textcolor{darkgreen}{\textbf{Constraint learning}}: We use approximate normal space data to learn a constraint distance pseudometric $d_\mathcal{M}$, which we use to reduce the constraint input space. We train the constraint $\mathcal{\hat M}$ using input-reduced offline and online data, paired with the normal space data. \textcolor{blue}{\textbf{Evaluation}}: For prediction, we evaluate $\hat f(x,u)$ and project it onto $\mathcal{\hat M}$.}
    \label{fig:method}
    \vspace{-20pt}
\end{figure}

Our method is shown in Fig. \ref{fig:method}. We first learn nominal GP dynamics, where the GP input space is sparsified by a learned pseudometric (Sec. \ref{sec:sparse_dyn}). Next, we learn a sparsified GP for the constraints (Sec. \ref{sec:constraint_learning}), using approximate data (Sec. \ref{sec:approx_constraint}). Finally, we predict by projecting the nominal dynamics onto the learned constraint (Sec. \ref{sec:projection}).



\subsection{Learning sparse dynamics}\label{sec:sparse_dyn}

We describe a framework for obtaining a pseudometric via contrastive learning (Sec. \ref{sec:dist1}), how to specialize the method to identify the sparsity pattern of the dynamics (Sec. \ref{sec:dynamics_dist}), and how we use this knowledge to train dynamics which are sparse in their input space (Sec. \ref{sec:dynamics_gp}).

\subsubsection{Contrastive learning of pseudometrics}\label{sec:dist1}

We discuss a framework for learning pseudometrics, and specialize it in Sec. \ref{sec:dynamics_dist} and \ref{sec:constraint_dist} to dynamics and constraint learning.
Pseudometrics are distance metrics without the restriction that unique elements of the input space have nonzero distance, i.e., for all $z$, $z'$, $z'' \in \Z$, a pseudometric $d: \Z\times\Z \rightarrow \mathbb{R}_{\ge 0}$ satisfies: 1) $d(z,z) = 0$, 2) $d(z,z') = d(z',z)$, and 3) $d(z,z'') \le d(z,z') + d(z',z'')$ \cite{kelley1975general}. For a generic dataset $D = \{z_i\}_{i=1}^Z$, $z_i \in \Z$, we parameterize a pseudometric $d_\theta: \Z \times \Z \rightarrow \mathbb{R}_{\ge 0}$ with  parameters $P_\theta \in \mathbb{S}_{+}^{(\nx+\na) \times (\nx+\na)}$, where $\mathbb{S}_{+}$ is the space of positive semi-definite matrices, 
\begin{equation}\label{eqn:metric_learned}
    d_\theta(z_1, z_2) = (z_1 - z_2)^\top \Dist_\theta (z_1 - z_2),
\end{equation}
In general, we can search over input-dependent metrics (i.e., $\Dist_\theta$ is a function of $z$); for simplicity, we restrict attention to the case where $\Dist_\theta$ is constant.
To learn the pseudometric $d_\theta(\cdot,\cdot)$, we use contrastive learning \cite{DBLP:conf/cvpr/ChopraHL05}. In the context of metric learning, contrastive learning takes positive and negative pairs  as inputs.  Here, positive pairs are close in the output space and are above a distance threshold in the input space, and negative pairs are all others (see Sec. \ref{sec:dynamics_dist}).  Contrastive learning encourages $d_\theta(\cdot,\cdot)$ to evaluate to small (large) values for positive (negative) pairs. We specifically use the popular InfoNCE loss \cite{DBLP:journals/corr/abs-1807-03748}, defined as
\begin{equation}\label{eqn:contrastive}\small
    \mathcal{L}_\textrm{con} = -\mathbb{E}_{z\sim D}\Bigg[\log\frac{\exp(\textrm{sim}(z,z^p))}{\sum_{k=1}^{N_b} \exp(\textrm{sim}(z, z_k^n))}\Bigg]
\end{equation}
where $(z,z^p)$ is a positive pair, $(z,z_k^n)$ is one of $N_b$ negative pairs, and $\textrm{sim}(\cdot,\cdot)$ is a similarity metric (we use cosine similarity, commonly used in contrastive learning \cite{DBLP:conf/icml/ChenK0H20}, which is a discriminative distance metric in high-dimensional spaces),

\vspace{-10pt}
\begin{equation}\small
    \textrm{sim}(z,z') = \frac{z^T \Dist_\theta z'}{\sqrt{z^\top \Dist_\theta z} \sqrt{{z'}^\top \Dist_\theta z'}}
\end{equation}
Then, we can train such a pseudometric by minimizing $\mathcal{L}_\textrm{con}$.

\subsubsection{Learning a pseudometric for sparsifying the dynamics}\label{sec:dynamics_dist}

We employ the framework of Sec. {\ref{sec:dist1}} to learn a pseudometric that extracts the sparsity pattern of \eqref{eqn:dyn} from data. Consider a pseudometric that assigns distances between $(x,u)$ and $(x',u')$ to be small (large) if they have (dis)similar derivatives $f(x,u)$ and $f(x',u')$. Then, if the dynamics are invariant along any directions $v$, i.e., $f(x,u) = f(x+\alpha v,u)$, $\forall \alpha \in \mathbb{R}$, an ideal pseudometric would assign $d((x,u),(x+\alpha v,u)) = 0$. For a pseudometric of the form \eqref{eqn:metric_learned}, we can recover these directions $v$ as the eigenvectors of $\Dist_\theta$ corresponding to zero eigenvalues. When $\Dist_\theta$ is diagonal (as in the results), each eigenvector corresponds to a specific dimension of the input. We discuss how to use this knowledge for dynamics learning in Sec. \ref{sec:dynamics_gp}. To actually obtain such a pseudometric, we can leverage the contrastive learning framework of Sec. \ref{sec:dist1}. In particular, we select the input $z$ as the state-control space $(x,u)$. We define the positive pair corresponding to $(x_i,u_i)$ as $(x_{i^*},u_{i^*})$, where $i^* = \arg\min_{j\in \D} \Vert f(x_i, u_i) - f(x_j, u_j)\Vert$ such that $\Vert (x_i, u_i) - (x_{i*}, u_{i*})\Vert \ge \epsilon_1$. 
Negative pairs are selected as the other non-minimum pairs in the batch. Training the dynamics 
pseudometric $d_f$ is done by minimizing \eqref{eqn:contrastive}.

\subsubsection{Learning the sparse dynamics model}\label{sec:dynamics_gp}

 By following Sec. \ref{sec:dynamics_dist}, we can obtain a pseudometric that sets distances along invariant directions to zero. For diagonal $\Dist_\theta$, we can drop out input dimensions corresponding to diagonal entries equal to zero when learning the dynamics. We refer to the reduced input as $(x_\textrm{red}, u)$; note that we only reduce the input, not the output, which remains in $\mathbb{R}^\nx$. If we have noisy data or an imperfect pseudometric, we can drop out inputs if the diagonal entries in $\Dist_\theta$ lie below some threshold. 
Intuitively, this performs principal component analysis on the input data, removing input ``features" that are not predictive for the dynamics. If the removed states globally do not affect the dynamics (reasonable for smooth systems with enough data), we can expect the reduced model to generalize better in OOD scenarios, as it has fewer inputs upon which to be OOD. We show in Sec. \ref{sec:results} that this reduction improves predictions in practice. Similar logic applies for non-diagonal $\Dist_\theta$, e.g., we can eigen-decompose $\Dist_\theta$, rotate the data using the eigen-basis, and remove dimensions of the rotated data with small eigenvalues.
We learn the sparsified dynamics with a GP on the reduced space, where the data $(x_i, u_i, f(x_i, u_i))$ is used to train an IGP by maximizing log-likelihood, 
giving the learned model
\begin{equation}\label{eqn:learned_dynamics}
    \dot x = \hat f(x_\textrm{red}, u).
\end{equation}
We make predictions via the posterior mean of the GP. 


 
\subsection{Learning the constraint manifold}\label{sec:constraint_learning}

We describe our method for learning the constraint manifold \eqref{eqn:manifold}. First, we detail our constraint formulation (Sec. \ref{sec:constraint_formulation}) and discuss challenges in identifying the constraints from finite data. We discuss how to overcome these challenges using synthetic data (Sec. \ref{sec:approx_constraint}), and how we can improve generalization of the learned constraint by learning a constraint pseudometric (Sec. \ref{sec:constraint_dist}). Finally, given this, we show how to learn the constraint manifold (Sec. \ref{sec:constraint_gp}).

\subsubsection{Constraint formulation}\label{sec:constraint_formulation}

As discussed in Sec. \ref{sec:prelims}, we assume the unknown constraint can be represented as in \eqref{eqn:pfaffian_matrix} as an affine function of $\dot{x}$.
Thus, the constraint learning problem reduces to learning $\Gamma: \mathcal{X} \rightarrow \mathbb{R}^{\nc\times \nx+1}$. For systems of interest, $\nc < \nx$, and controls can be taken to generate state derivatives $\dot x$ which satisfy \eqref{eqn:pfaffian_matrix}. 
  In principle, to recover $\Gamma(x)$, we can apply different control actions $u_i$ at a fixed state $x$ until we obtain $\nx-\nc$ linearly independent augmented derivatives $\{[f(x,u_i)^\top\ 1]\}_{i=1}^{\nx-\nc}$. For concreteness, we denote the concatenation of these vectors as $\dot X(x) \in \mathbb{R}^{(\nx-\nc) \times (n+1)}$:
  
  \vspace{-7pt}
  \begin{equation}\label{eqn:Xdot_matrix}\small
    \dot X(x) = \begin{bmatrix} f(x, u_1) & \cdots & f(x, u_{\nx-\nc}) \\ 1 & \cdots & 1 \end{bmatrix}^\top.
  \end{equation}
  \vspace{-11pt}
  
  Then, we can compute a basis for the null space of $\dot X(x)$ to recover $ G({x})$, $g(x)$. However, it is difficult to reset the state of a robot with non-trivial dynamics in a way that different actions can be taken at the exact same $x$. Thus, we propose a method to approximate $G(x)$ and $g(x)$ using synthetic data. 

\subsubsection{Approximating the normal space}\label{sec:approx_constraint}

To estimate $\dot X(x)$, we use the learned $\hat f$ as a proxy for $f$ to compute a synthetic Learned Multi-Action (LMA) dataset, for $K \ge \nx-\nc$:

\vspace{-7pt}
\begin{equation}\label{eqn:Xdot_matrix_lma}\small
      \dot X_\textrm{LMA}(x) = \begin{bmatrix} \hat f(x, u_1) & \cdots & \hat f(x, u_K) \\ 1 & \cdots & 1 \end{bmatrix}^\top.
\end{equation}
\vspace{-11pt}

Given an approximation of $\dot X(x)$, we can compute its null space to approximate $\Gamma(x)$. We do this by applying the singular value decomposition (SVD) to $\dot X_\textrm{LMA}(x)$, 
\begin{equation}\small
    \hspace{-5pt}\dot{X}_\textrm{LMA}(x) = U\hat{\Sigma} V^\top = [U_1\ U_2] \begin{bmatrix}
        \hat{\Sigma}_1& \mathbf{0}\\
        \mathbf{0} & \hat{\Sigma}_2
    \end{bmatrix}
     \begin{bmatrix}
            V_1^\top \\V_2^\top
        \end{bmatrix},
\label{eqn:svd}
\end{equation}
where $\Sigma_2$ contains all singular values less than a threshold $\epsilon$, and $V_2$ is an approximate basis for the null space of $\dot X(x)$, i.e., $\Gamma(x) = V_2^\top$. In general, $\epsilon$ should be positive to handle any approximation error in $\dot X(x)$ on $\Gamma(x)$, and can also be used to identify the number of constraints $\nc$. 
However, since the SVD is not unique \cite{article2}, directly using $V_2$ to represent $\Gamma(x)$ can cause challenges for learning, as $V_2$ may not change smoothly with $x$. While in general, no smoothly-varying basis exists \cite{DBLP:journals/mp/ByrdS86}, the following method is empirically effective. We first convert $V_2^\top $ to reduced row echelon form (rref); and define  $\Tilde{\Gamma}^{(i)}(x)$ as the $i$th row of $\textrm{rref}(V_2^\top(x))$, which smoothly standardizes the basis up to a sign-change.  
To keep signs consistent, at a given $x$, we find the nearest datapoint $x_\textrm{ref}$, and negate the vectors in $\textrm{rref}(V_2^\top(x))$ if their dot products with $\textrm{rref}(V_2^\top(x_\textrm{ref}))$ lie below zero. We call this approximate normal space basis $\Gamma_{\textrm{approx}}(x)$, where the $i$th row is:

\vspace{-13pt}
\begin{equation}\label{eqn:approx_gamma}
    \Gamma^{(i)}_{\textrm{approx}}(x) \doteq
\begin{cases}
    \Tilde{\Gamma}^{(i)}(x),& \text{if }  \Tilde{\Gamma}^{(i)}(x) \Tilde{\Gamma}^{(i)}(x_\textrm{ref})^T \ge0\\
    -\Tilde{\Gamma}^{(i)}(x),              & \text{if }  \Tilde{\Gamma}^{(i)}(x) \Tilde{\Gamma}^{(i)}(x_\textrm{ref})^T <0
\end{cases}
\end{equation}
\vspace{-7pt}

We can then form a dataset $\{x_i, \Gamma_\textrm{approx}(x_i)\}_{i=1}^{|\D|}$ to train a GP constraint in Sec. \ref{sec:constraint_gp}. 
Before training, we discuss how to learn a pseudometric (similar to Sec. \ref{sec:dynamics_dist}) to reduce the input space of the constraint, improving generalization.

\subsubsection{Training the constraint distance pseudometric}\label{sec:constraint_dist}

For the learned constraint to be useful in aiding OOD generalization of the dynamics, the constraints must also be accurate OOD. To improve generalization as is done in Sec. \ref{sec:dynamics_dist}, we learn $\nc$ distance pseudometrics, one for each of the $\nc$ rows of $\Gamma(x)$, with the aim of reducing the input space when learning the constraint. Here, we take the contrastive learning framework of Sec. \ref{sec:dist1}, and let $z = x$. Denote the distance pseudometric for to the $i$th row of $\Gamma(x)$, $\Gamma^{(i)}(x)$, as 
\begin{equation}\label{eqn:metric_learned_constraint}
    d_{\mathcal{M}^{(i)}}(x, x') = (x - x')^\top \Dist_{\mathcal{M}^{(i)}} (x - x'),
\end{equation}
For each datapoint $x_i$, we select its positive pair as the point $x_i^p$ in $\D$ minimizing $\Vert \Gamma_\textrm{approx}^{(i)}(x_i) - \Gamma_\textrm{approx}^{(i)}(x_i^p)\Vert$  satisfied  $\Vert x_i - x^p_i\Vert \ge \epsilon_2$. Negative pairs are selected as the remaining non-minimum states in the batch. We train each pseudometric by minimizing \eqref{eqn:contrastive}. 

\subsubsection{Learning the constraint on reduced input space}\label{sec:constraint_gp}

 Similar to Sec. \ref{sec:dynamics_gp}, for each row of $\Gamma(x)$, we drop out the inputs that correspond to a zero eigenvalue. We refer to the reduced input state for row $i$ as $x_\textrm{red}^i$, $i = 1,\ldots \nc$. We model each entry of the normal space matrix using an IGP, where each is a function of its corresponding reduced input space. Specifically, we follow the framework of Sec. \ref{sec:prelims}, where the training data and labels is comprised of states $x_i$ and approximate normal space bases $\Gamma_\textrm{approx}(x_i)$ (see \eqref{eqn:approx_gamma}), respectively, where $x_i$ are drawn from the dataset $\D$ described in Sec. \ref{sec:prelims}. Using this data, we train an IGP by maximizing log-likelihood.
 We call this learned constraint $\hat \Gamma(x) = [\hat G(x)\ \ \hat g(x)]$, and the corresponding manifold,
 \begin{equation}\label{eqn:learned_constraint}
    \mathcal{\hat M} \doteq \{(x,\dot x) \in \mathcal{X}\times T_x(\mathcal{X}) \mid \hat G(x) \dot x + \hat g(x) = 0\},
\end{equation}
which is evaluated via the trained GP's posterior mean.

\subsection{Generating predictions at evaluation time} \label{sec:projection}

At runtime, we wish to make predictions with the learned dynamics $\dot x_\textrm{pred} = \hat f(x,u)$ \eqref{eqn:learned_dynamics} that conform with the learned constraint $\mathcal{\hat M}$ \eqref{eqn:learned_constraint}. Since for a fixed $x$, the constraint \eqref{eqn:learned_constraint} is affine in $\dot x$, we can \textit{project} the prediction $\dot x_\textrm{pred}$ from the nominal dynamics \eqref{eqn:learned_dynamics} to be consistent with $\mathcal{\hat M}$ by solving:
\begin{equation}\label{eqn:projection}\small
    \begin{array}{rl}
         \textrm{proj}_\mathcal{M}(\dot x_\textrm{pred}) \doteq \underset{\dot x}{\textrm{argmin}} & \Vert \dot x - \dot x_\textrm{pred} \Vert_2^2 \\
        \textrm{subject to} & \hat G(x)\dot x + \hat g(x) = 0,
    \end{array}
\end{equation}
which can be efficiently done online by solving a quadratic program (QP). We use \eqref{eqn:projection} for prediction in Sec. \ref{sec:results}.

\section{Results}\label{sec:results}

We evaluate our method on a simulated unicycle, physical differential-drive robot (DDR), and simulated quadrotor. First, we describe our baselines, ablations, method variants, and test settings. Our baseline is a standard GP trained on $\D$ (``GP" in Tab. \ref{table: unicycle}-\ref{table:quadrotor}) (i.e., no sparsity or constraint). We refer to our sparsified GP (i.e., with dropped inputs) as ``DGP". Our method, which also projects onto the constraint learned via \eqref{eqn:svd}, is ``DGP + Approx.", the ideal version of our full method (which projects onto the constraint learned via $\dot X(x)$ \eqref{eqn:Xdot_matrix}) is ``DGP + Ideal", and an ablation which sparsifies but does not project is ``DGP + None". ``ID" evaluates each method on in-distribution (ID) test data; ``OOD" on OOD test data. We consider test data as ID if it lies in the support of the distribution that the offline data was sampled from, and OOD otherwise. For the OOD experiments, we evaluate all methods when given both offline and online data, i.e., where $\D = \D_\textrm{off} \cup \D_\textrm{on}$. Finally, to show robustness to noise, we evaluate on OOD data where $\D_\textrm{off}$ is noiseless but $\D_\textrm{on}$ is corrupted with Gaussian measurement noise $N(0,\eta I)$. We test with $\eta = 0.05$ and $\eta = 0.1$. 
All dynamics GPs are trained using a radial basis function (RBF) kernel. For the corrupted dataset we also employed low-pass filtering to enhance robustness against high-frequency noise. We simulate with timestep $\Delta T = 0.01$ for the quadrotor and $\Delta T = 0.1$ otherwise, and report cumulative prediction error, 
\begin{equation}\label{eqn:loss}
        \mathcal{L}_\textrm{err} = \textstyle\sum_{t=0}^{T}  \Vert x_t^{\textrm{true}} - x_t^{\textrm{pred}} \Vert_2.
\end{equation}


\noindent \textbf{Nonholonomic unicycle ($x\in \mathbb{R}^3$, $u\in \mathbb{R}^2$)}: The dynamics are
\begin{equation}\footnotesize
\begin{bmatrix}
\dot{p}_x \\
\dot{p}_y \\
\dot{\theta}
\end{bmatrix} =
\begin{bmatrix}
0& \cos\theta \\
0& \sin\theta \\
1&0
\end{bmatrix} 
\begin{bmatrix}
\omega\\
v\\
\end{bmatrix},
\label{eqn:car}
\end{equation}
where $u = (\omega, v)$ are controls (angular, linear velocity), $p_x$, $p_y$ are the position, and $\theta$ is the heading. 
The unicycle satisfies the constraint $-\dot p_x\sin(\theta) + \dot p_y \cos(\theta) = 0$, i.e., $G(x) = [-\sin(\theta)\ \ \cos(\theta)\ \ 0]$ and $\nc=1$. 
We learn a dynamics model with input  $[x, \hspace{1mm} u] =[p_x,p_y,\theta,\omega,v]^\top \in \mathbb{R}^5$ and output $y = [\dot{p}_x,\dot{p}_y,\dot{\theta}]^\top \in \mathbb{R}^3$. As the dynamics are only affected by $\theta$ and $u$, we learn a dynamics pseudometric \eqref{eqn:metric_learned} with zero $\epsilon_1$, giving  $\Dist_f = \textrm{diag}([0, 0, 1.15, 1.66, 1.63])$ (cf. Sec. \ref{sec:dynamics_dist}). We use $\Dist_f$ to remove $p_x$ and $p_y$ from the input when learning the dynamics \eqref{eqn:learned_dynamics} . We also learn the constraint pseudometric \eqref{eqn:metric_learned_constraint} with zero $\epsilon_2$ , where $\Dist_\mathcal{M} = \textrm{diag}([0, 0., 1.59])$. Thus, we remove $p_x$ and $p_y$ from the input when learning the constraint \eqref{eqn:learned_constraint}, for which we use an RBF kernel. 
 The offline dataset $\D_\textrm{off}$ contains $20$ trajectories, generated by applying $25$ random controls in $[\omega,v] \in [-0.2,0.2] \times [0, 1.5]$, starting from initial conditions in $[p_x,p_y,\theta] \in [-10, 10]^2 \times[-\frac{\pi}{3}, \frac{\pi}{3}]$. For the online dataset $\D_\textrm{on}$, we sample an initial state from $[p_x,p_y,\theta]\in [15, 20]^2\times[\frac{\pi}{3}, \frac{\pi}{2}]$, and roll out by sampling $5$ controls in $[\omega,v] \in [-0.2,0.2] \times [0, 1.5]$, and holding each for $5$ steps, giving 25 total datapoints. Finally, starting from the end of the online trajectory, we test the $20\times 5 = 100$ multi-step prediction error ($20$ actions, sampled from $[\omega,v] \in [-0.2,0.2] \times [0, 1.5]$, each held for $5$ steps), and generate 5 such trajectories for evaluation. $\D_\textrm{on}$ is generated 10 times, each starting from a unique initial state; test data is also regenerated. See Fig. \ref{fig:data_unicycle} (left) for $\D_\textrm{off}$, one of $\D_\textrm{on}$, and the associated $\D_\textrm{test}$. The prediction accuracies, averaged over all online datasets and test trajectories, are in Table \ref{table: unicycle}. 



\begin{table}
\begin{center}
\begin{adjustbox}{width=0.9\columnwidth,center}

\begin{tabular}{|c|c||c|c|c|}
\hline
\multicolumn{2}{|c||}{}                     & None & Ideal & Approx. \\ \hline
\multicolumn{1}{|c|}{\multirow{2}{*}{ID}}& GP &  4.79  $\pm$ 4.17  & 3.63  $\pm $ 2.53  &  4.05 $\pm$ 3.40 \\ \cline{2-5} 
\multicolumn{1}{|c|}{}                  &  DGP& 2.96 $\pm$ 2.99 & 2.56  $\pm$ 2.21  &  3.03 $\pm$ 3.19 \\ \hline

\multicolumn{1}{|c|}{\multirow{2}{*}{\shortstack{OOD,\\ $\eta = 0.0$}}}&
GP & 9.08 $\pm$ 1.14 &
7.70 $\pm$ 1.01&
7.82 $\pm$ 1.25\\ \cline{2-5} 
\multicolumn{1}{|c|}{} &  DGP& 3.94 $\pm$ 0.48 & 
3.51 $\pm$ 0.44
& 3.76 $\pm$ 0.71  \\ \hline 
\multicolumn{1}{|c|}{\multirow{2}{*}{\shortstack{OOD,\\ $\eta = .05$}}}&
GP & 43.58 $\pm$ 25.27 &
41.66 $\pm$ 26.47&
43.42 $\pm$ 24.80\\ \cline{2-5} 
\multicolumn{1}{|c|}{}                  &  DGP& 35.52 $\pm$ 25.10 & 
33.46 $\pm$ 25.90
& 35.88 $\pm$ 25.03  \\ \hline
\multicolumn{1}{|c|}{\multirow{2}{*}{\shortstack{OOD,\\ $\eta = .10$}}}&
GP & 56.05 $\pm$ 37.13 &
55.41 $\pm$ 35.85 &
54.05$\pm$ 31.73\\ \cline{2-5} 
\multicolumn{1}{|c|}{}                  &  DGP& 45.44 $\pm$ 30.51 & 
44.34 $\pm$ 29.79
& 45.29 $\pm$ 28.87  \\ \hline
\end{tabular}
\end{adjustbox}
\end{center}\vspace{-8pt}
\caption{100-step prediction error (Unicycle). Data: $\D_\textrm{on} \cup \D_\textrm{off}$.}\vspace{-33pt}
\label{table: unicycle}
\end{table}

\begin{figure}
    \centering\vspace{-5pt}
    \includegraphics[width=\linewidth]{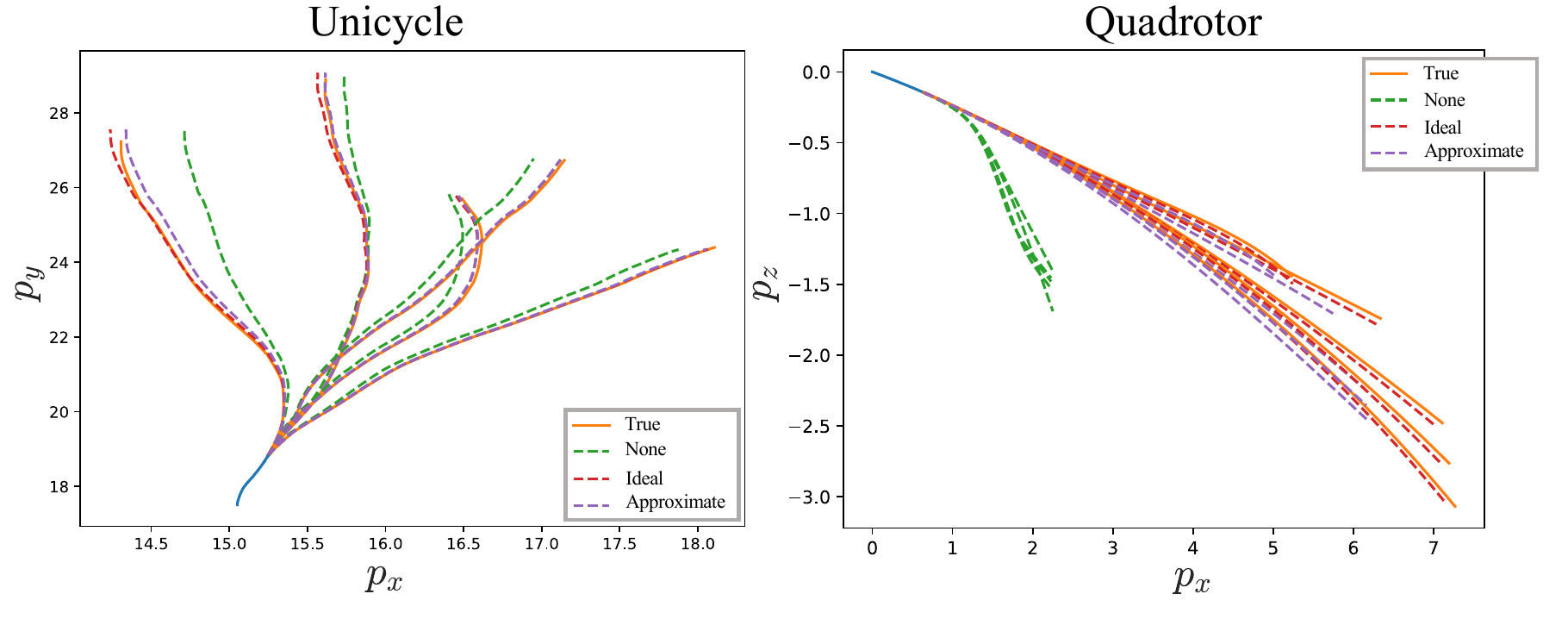}\vspace{-12pt}
    \caption{Some predicted trajectories for the unicycle (\textbf{left}) and quadrotor (\textbf{right}) on OOD data. The given (noiseless) online data $\D_\textrm{on}$ is in blue. ``True": ground truth trajectory. The predicted trajectories are: ``None": GP, no projection. ``Ideal": DGP, with projection on ideal learned constraint. ``Approximate": DGP, with projection on approximate learned constraint.}
    \label{fig:unicycle_traject}
    \vspace{-8pt}
\end{figure}

When given both $\D_\textrm{off}$ and $\D_\textrm{on}$ to train the dynamics and constraint, we see for ID predictions that all method variants have relatively similar performance, with some small improvements from DGP and constraint projection. This is not surprising, as the baseline GP is already quite accurate and confirms that projection or sparsification does not hurt performance where data is plentiful. For OOD predictions, the sparse DGP is uniformly more accurate than the GP baseline, for all the noise levels tested. When further combining DGP with projection, ``DGP + Ideal" performs best on the noiseless case. There is minor degradation for ``DGP + Approx." due to the error in the training data, but it still is more accurate than the variants with no projection. With larger magnitudes of measurement noise, DGP still performs uniformly better than the GP variants, while the improvement due to projection shrinks for higher noise. This is likely due to the long prediction horizon in this problem, which makes it critical to estimate the constraint with high accuracy, which is more difficult to achieve with noisy data, whereas the sparsity pattern can be more easily estimated. We visualize the true and predicted trajectories in Fig. \ref{fig:unicycle_traject} (left). Overall, these results suggest that both sparsifying and projection can improve prediction accuracy, in spite of noisy data.

\begin{figure}
    \centering\vspace{-0pt}
    \includegraphics[width=\linewidth]{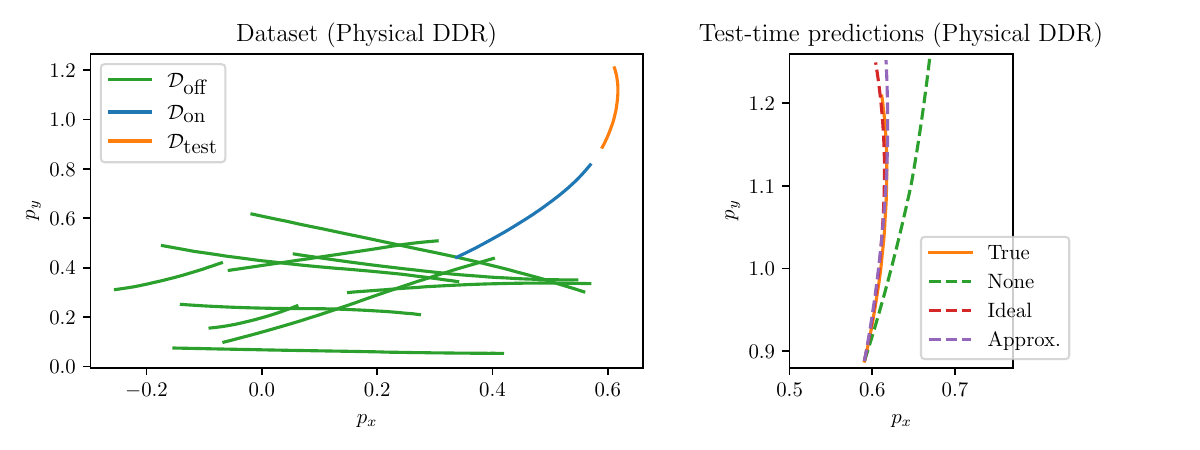}\vspace{-12pt}
    \caption{Hardware diff-drive robot: (\textbf{left}) offline data and one of the online and associated test datasets; (\textbf{right}) examples of predicted trajectories.}
    \label{fig:ddr_traject}
    \vspace{-20pt}
\end{figure}
 
\noindent \textbf{Hardware DDR}: To show our method's robustness to non-idealized constraints, we evaluate on a physical DDR (cf. Fig. \ref{fig:manifold}, right), where we assume the same state/control space as the unicycle and data is collected using a Vicon mocap system. We learn a dynamics pseudometric $\Dist_f = \textrm{diag}([0,0.04,1.72,0.43,0.44])$ and constraint pseudometric $\Dist_\mathcal{M} = \textrm{diag}([0,0,1.60])$ with $\epsilon_1$ and  $\epsilon_2$ set to 0.5, so we drop out $p_x$, $p_y$ from the input in both cases as their diagonal values fall below a threshold. We use a cosine kernel when learning the constraint. The offline dataset $\D_\textrm{off}$ contains 10 trajectories, generated by applying 2 random controls from $(w, v) \in [-0.2, 0.2] \times [0, 0.2]$ and with initial conditions from $(p_x, p_y, \theta) \in [-0.2, 0.6]\times [0.05, 0.6] \times [-0.6, 0.8]$. For the online dataset $\D_\textrm{on}$, we sample from the same $u$ range, but with initial conditions from $(p_x, p_y, \theta) \in [-0.4, 0.6]\times [0.4, 1.2] \times [-1.5, 1.7]$. We generate $\D_\textrm{on}$ 8 times, and generate 1 associated $\D_\textrm{test}$ per online dataset. See Fig. \ref{fig:ddr_traject} for the datasets and predicted trajectories, and Tab. \ref{table: OOD_unicycle_real} for prediction accuracies, averaged over all online datasets and test trajectories. We use both $\D_\textrm{off}$ and $\D_\textrm{on}$ to train the dynamics and constraint. For the ID range, all methods are comparable; this is not surprising, as the baseline model is already quite accurate, moreover supporting that our method does not degrade predictions near data. With evaluation in the OOD range, DGP is uniformly more accurate than the GP baseline (cf. Tab. \ref{table: OOD_unicycle_real}), again supporting that sparsity improves OOD accuracy. Constraint projection further improves prediction accuracy, where in Tab. \ref{table: OOD_unicycle_real}, the ``Ideal" column is computed by projecting onto the unicycle constraint $-\dot p_x \sin(\theta) + \dot p_y \cos(\theta) = 0$. Crucially, as the hardware can violate this ideal constraint, this shrinks the accuracy gap between the ideal and approximate projection. Finally, as more noise is injected into $\D_\textrm{on}$, all methods degrade, but DGP with projection (our full method) remains more accurate than baselines without sparsity or projection, and for $\eta = 0.10$ slightly outperforms the ideal case. Overall, this suggests that in spite of the approximate data, constraint-learning can be valuable compared to pre-specifying constraints that may fail to hold in reality. 


\begin{table}
\begin{center}
\begin{adjustbox}{width=0.75\columnwidth,center}
\begin{tabular}{|c|c||c|c|c|}
\hline
\multicolumn{2}{|c||}{}                     & None & Ideal & Approx. \\ \hline

\multicolumn{1}{|c|}{\multirow{2}{*}{ID}}& GP & 0.60  $\pm$ 8e-4  & 0.61  $\pm $ 1e-7  &  0.61 $\pm$ 8e-4 \\ \cline{2-5} 
\multicolumn{1}{|c|}{}                  &  DGP& 0.50 $\pm$ 5e-3 & 0.52  $\pm$ 6e-4  &  0.51 $\pm$ 4e-3 \\ \hline

\multicolumn{1}{|c|}{\multirow{2}{*}{\shortstack{OOD,\\ $\eta = 0.0$}}}&
GP & 1.33 $\pm$ 0.62 &
1.03 $\pm$ 0.37&
1.04 $\pm$ 0.37\\ \cline{2-5} 
\multicolumn{1}{|c|}{}                  &  DGP& 0.75 $\pm$ 0.32 & 
0.68 $\pm$ 0.27
& 0.70 $\pm$ 0.25  \\ \hline

\multicolumn{1}{|c|}{\multirow{2}{*}{\shortstack{OOD,\\ $\eta = .05$}}}&
GP & 1.68  $\pm$ 0.93 &
0.90 $\pm$ 0.35&
0.95 $\pm$ 0.33\\ \cline{2-5} 
\multicolumn{1}{|c|}{}                  &  DGP& 1.58 $\pm$ 0.78 & 
0.83 $\pm$ 0.27
& 0.90 $\pm$ 0.25  \\ \hline
\multicolumn{1}{|c|}{\multirow{2}{*}{\shortstack{OOD,\\ $\eta = .10$}}}&
GP & 2.15 $\pm$ 0.88 &
1.42 $\pm$ 0.57&
1.37 $\pm$ 0.46\\ \cline{2-5} 
\multicolumn{1}{|c|}{}                  &  DGP& 1.85 $\pm$ 0.57 & 
1.38 $\pm$ 0.42
& 1.33 $\pm$ 0.34  \\ \hline
\end{tabular}
\end{adjustbox}
\end{center}\vspace{-8pt}
\caption{40-step prediction error (Hardware DDR). Data: $\D_\textrm{on} \cup \D_\textrm{off}$.}\vspace{-26pt}
\label{table: OOD_unicycle_real}
\end{table}


\noindent \textbf{Planar quadrotor ($x\in \mathbb{R}^6$, $u\in \mathbb{R}^2$)}: The quadrotor satisfies
\begin{equation}\footnotesize
\begin{bmatrix}
\dot{p_x}\\ \dot{p_z} \\ \dot{\phi} \\ \dot{v}_x \\ \dot{v}_z \\ \dot{\omega}_\phi
\end{bmatrix}  = 
\begin{bmatrix}
v_x\\ v_z \\ \omega_\phi \\ 0 \\ -g \\ 0
\end{bmatrix} + 
\begin{bmatrix}  0 & 0\\ 0 & 0 \\ 0 & 0  \\ -\frac{1}{m}\sin(\phi)& -\frac{1}{m}\sin(\phi) \\ \frac{1}{m}\cos(\phi)&\frac{1}{m}\cos(\phi)\\ \frac{1}{J} & -\frac{1}{J}
  \end{bmatrix}
  \begin{bmatrix}
u_1\\
u_2\\
\end{bmatrix} ,
\label{eqn:quadrotor}
\end{equation}
with positions, $p_x, p_z$, velocities $v_x, v_z$, orientation $\phi$, and angular velocity $\omega_\phi$, and $m=0.486$, $l=0.25$, and $J=0.0383$. There are $\nc=4$ constraints, analytically derived as:
\begin{equation}\footnotesize
    \Gamma(x) = \begin{bmatrix}
        1 & 0 & 0 & 0 & 0 & 0 & -v_x \\ 0 & 1 & 0 & 0 & 0 & 0 & -v_z \\ 0 & 0 & 1 & 0 & 0 & 0 & -\omega_\phi \\ 0 & 0 & 0 & 1 & \tan(\phi) & 0 & g\tan(\phi)
    \end{bmatrix}.
\end{equation}
We learn    $\Dist_f = \textrm{diag}([0, 0, 1.2, 1.4, \allowbreak 1.4, 1.3, 1.1, 1.0])$ with zero $\epsilon_1$. Thus, we remove $p_x$ and $p_z$ from the input when learning the dynamics \eqref{eqn:learned_dynamics}. We also learn $\nc$ constraint pseudometrics \eqref{eqn:metric_learned_constraint} with zero $\epsilon_2$, where $\Dist_{\mathcal{M}^{(i)}}$ recover the sparsity patterns of $x$ ($\Dist_{\mathcal{M}^{(i)}}, i = 1,...,4$, which only have nonzero entries on their 4th, 5th, 6th, and 3rd diagonals, respectively). We use this to drop out the associated inputs when learning the constraint \eqref{eqn:learned_constraint}, where we use a linear kernel.
The offline dataset $\D_\textrm{off}$ contains $5000$ trajectories, generated by applying $2$-step random $u$ from $[\frac{mg}{2}, 2.5]\times[\frac{mg}{2}, 2.5]$, with initial conditions sampled from $[p_x,p_y,\phi,v_x,v_z,\omega_\phi]\in [-1, 1]\times[-1, 1]\times[-\frac{\pi}{3}, \frac{\pi}{3}] \times [0,5] \times[-5,0]\times[-\frac{\pi}{3}, \frac{\pi}{3}] $. For $\D_\textrm{on}$, we sample $5$ control actions from the same set, each held for $10$ steps, which typically takes the system far from $\D_\textrm{off}$ (cf. Fig. \ref{fig:data_unicycle}). We test the $10\times 10 = 100$ multi-step prediction error ($10$ random actions held for $10$ steps each) on $5$ such trajectories (this is $\D_\textrm{test}$). We regenerate $\D_\textrm{on}$ and the associated $\D_\textrm{test}$ for $10$ initial conditions, and give statistics in Tab. \ref{table:quadrotor}.




\begin{table}
\begin{center}
\begin{adjustbox}{width=0.85\columnwidth,center}
\begin{tabular}{|c|c||c|c|c|}
\hline
\multicolumn{2}{|c||}{}              & None & Ideal & Approx. \\ \hline
\multicolumn{1}{|c|}{\multirow{2}{*}{ID}}& GP &  17.58	$\pm$	16.11     & 8.75	$\pm$	7.27 
& 10.75	$\pm$	9.99  \\ \cline{2-5} 
\multicolumn{1}{|c|}{}            &  DGP&  9.43	$\pm$	8.25      & 3.15	$\pm$	2.29 & 6.20	$\pm$	5.85 \\ \hline
\multicolumn{1}{|c|}{\multirow{2}{*}{\shortstack{OOD,\\ $\eta = 0.0$}}}&
GP& 87.91	$\pm$	77.34 & 35.21	$\pm$	29.30 & 71.64	$\pm$	69.15\\ \cline{2-5} 
\multicolumn{1}{|c|}{}                  &  DGP& 66.93	$\pm$	68.50 
& 23.79	$\pm$	28.19  
&  64.78	$\pm$	72.08 \\ \hline

\multicolumn{1}{|c|}{\multirow{2}{*}{\shortstack{OOD,\\ $\eta = .05$}}}&
GP & 187.58 $\pm$ 92.03 &
73.21 $\pm$ 34.43&
113.98 $\pm$ 78.24\\ \cline{2-5} 
\multicolumn{1}{|c|}{}                  &  DGP& 156.14 $\pm$ 97.32 & 
41.48 $\pm$ 37.08
& 85.36 $\pm$ 86.06  \\ \hline

\multicolumn{1}{|c|}{\multirow{2}{*}{\shortstack{OOD,\\ $\eta = .10$}}}&
GP & 216.14 $\pm$ 95.72 &
93.91 $\pm$ 51.90&
122.68 $\pm$ 73.83\\ \cline{2-5} 
\multicolumn{1}{|c|}{}                  &  DGP& 181.81 $\pm$ 104.44 & 
58.85 $\pm$ 58.95
& 90.15 $\pm$ 82.83  \\ \hline
\end{tabular}
\end{adjustbox}
\caption{100-step prediction error (Quadrotor). Data: $\D_\textrm{on} \cup \D_\textrm{off}$.}\vspace{-35pt}
\label{table:quadrotor}
\end{center}
\end{table}

In the ID case, even though the baseline is already quite accurate, projection onto the ideal and approximate constraints both improve performance $\approx$2-fold. Moreover, our approximate normal space training data does not overly degrade accuracy relative to the ideal case, as the ``Ideal" and ``Approx." variants achieve comparable accuracy. On the other hand, sparsification hurts accuracy slightly, likely since with more inputs, the model can overfit better ID. 

For the OOD case, we shrink the training range to be $[p_x,p_y,\phi,v_x,v_z,\omega_\phi]\in [-0.5, 0.5]\times[-0.5, 0.5]\times[-\frac{\pi}{4}, \frac{\pi}{4}] \times [0,2] \times[-2,0]\times[-\frac{\pi}{3}, \frac{\pi}{3}] $ and randomly sample initial conditions for the online data from $[p_x,p_y,\phi,v_x,v_z,\omega_\phi]\in [-1, 1]\times[-1, 1]\times[-\frac{\pi}{3}, \frac{\pi}{3}] \times [0,5] \times[-5,0]\times[-\frac{\pi}{3},\frac{\pi}{3}] $, such that the test data is OOD. In this case, both sparsification and constraint projection improves accuracy, with projection onto the ideal constraint improving accuracy $\approx$3-fold. Moreover, despite errors in the constraint training data, projection onto the approximate constraint still leads to improvement over the no-projection case. As increasing levels of noise are added to $\D_\textrm{on}$, as before, all methods degrade, but our approach (DGP+Approx.) retains an improvement over ``GP + None". In fact, the improvement that ``DGP+Approx" enables over ``DGP+None" increases for larger noise; we hypothesize this is because the nominal dynamics are much more inaccurate with high noise and thus benefit more from the approximate projection. We plot example rollouts for the OOD case in Fig. \ref{fig:unicycle_traject} (right); here, the test data is OOD in all state dimensions. Compared to the baseline GP (green), our method's predictions are more accurate. Overall, these results suggest that both sparsification and projection onto the learned constraint improve prediction accuracy, especially in OOD scenarios, and that the learned constraint generalizes better here than the learned dynamics.


\section{Discussion and Conclusion}

We present a method for improving the OOD accuracy of dynamics models from data. We do this by learning a pseudometric to uncover the sparsity in the data and by approximating the normal space with a GP to estimate the constraint manifold that the system must evolve on. 

While we our method outperforms baselines in our experiments, there are still limitations and future directions to our work. We wish to scale our approach to higher-dimensional systems, e.g., deformable objects. For such complex systems, it may be easier to find state/control-dependent pseudometrics, rather than the pseudometrics with constant $\Dist_\theta$ that we consider here.  
We also wish to explore identification of sparsity in the output space, which can simplify constraint learning (as $\Gamma(x)$ is often quite sparse), and to use our method in MPC. 

\bibliographystyle{IEEEtran}
\bibliography{IEEEabrv,Bibliography}

\end{document}